# Comparing Machine Learning Approaches for Table Recognition in Historical Register Books

Stéphane Clinchant
Hervé Déjean
Jean-Luc Meunier
Naver Labs Europe
Meylan, France
*firstname.lastname*@naverlabs.com

Eva Lang
Bischöfliches Ordinariat Passau
Archiv des Bistums Passau
Passau, Germany
eva.lang@bistum-passau.de

Florian Kleber
TU Wien
Computer Vision Lab
Vienna, Austria
kleber@cvl.tuwien.ac.at

*Abstract*—We present in this paper experiments on Table Recognition in hand-written registry books. We first explain how the problem of row and column detection is modelled, and then compare two Machine Learning approaches (Conditional Random Field and Graph Convolutional Network) for detecting these table elements. Evaluation was conducted on death records provided by the Archive of the Diocese of Passau. Both methods show similar results, a 89 F1 score, a quality which allows for Information Extraction. Software and dataset are open source/data.

## I. Introduction

Over the last decades, numerous digitization campaigns have started to provide online access to a large volume of archival documents. For documents such as registry or census books, a vast majority are not reports or narratives, but tables, which can be either fully drawn and written by hand or use printed forms. Today, archival institutions are still facing major issues when trying to perform Information Extraction on these handwritten archival documents.

Among many other difficulties faced when designing a full Information Extraction workflow, we address in this paper the problem of table understanding: recognizing the structural organization of tables to extract data [1]. While in the considered documents (register books, see description Section 2), the vertical structures are relatively simple (non-hierarchical table composed of roughly 7-10 columns), the segmentation into rows turns out to be more challenging. A register book usually is filled in by several writers, each of which can have their own style to lay out the record information. We propose in this article to investigate the use of two Machine Learning methods in order to organize the table content into rows. We formulate the problem as a graph labelling problem (each textual element is a node of the graph, and edges correspond to neighbourhood relations between textual elements) and use two graph-based methods: the first method relies on graph Conditional Random Fields (gCRF) [19], while the second method is based on Graph Convolutional Networks (GCN) [15].

A comprehensive presentation about Table Understanding can be found in [1]. In terms of archival documents, the most relevant work has been conducted by the Intuidoc IRISA team [2]. They use an approach based on a table description language having a logical and a physical part which have to be designed by a user. From the image, they extract a set of line segments and match the image information with column and row information contained in the table description. [3] presents a full workflow for Information Extraction from French censuses, however, in their collection records are always laid out in a single table row.

A more related work in terms of method is [4], which applies CRF on top of two Multi-Layer-Perceptron classifiers in order to categorize handwritten text lines as member of a table or not. In this approach, they only used text line information as no graphical lines (used as column and/or row separator) are used in this collection of chemistry documents. [5] also uses CRF to locate table and tag constituent text lines, "*but knows nothing of columns, does not segment individual table cells horizontally or vertically, and does not tag individual cells as being data or header.*". They describe in their Future Work Section "*a model which is connected in a grid-pattern instead of a linear-chain.*". Our gCRF is in spirit similar to their future (envisioned?) work.

Recently, [6] proposes DeepDeSRT, a combination of two deep learning systems, one for table detection, and a second for table understanding. The latter is a fully-connected network, where some pre-processing consists in stretching the images horizontally and vertically in order to improve the row and column identification. Besides this method, the paper argues that the image-based solution (converting any document, esp. PDF into image format) allows for generic and robust generic solutions. In our approach, we chose the opposite way: converting first images into 'PDF-like' documents. This choice is possible due to the quality of the 'pre-processing' tools used for converting image into structured objects (graphs).

Section II presents our dataset and our general use-case, explaining the Table Understanding problem for this dataset in detail. Section III first describes the way we formulate our main problem, row detection, and then details both Machine Learning approaches (gCRF and GCN) use to solve it. Section IV presents the various experiments and evaluations done, showing that both approaches perform well for this task, before we finally discuss current and future extensions.

## II. Dataset and Task

In this section, we first explain our dataset, and the general Information Extraction use-case we want to address, as well as the pre-processing steps applied before the table row detection task, the main task described in this paper.

*A. Data Description and Use-Case*

The ABP_S_1847-1878 dataset contains information about the parishioners who died within the geographic boundaries of the various parishes of the Diocese of Passau between the years 1847 and 1878. The dataset holds a total of 26,579 scanned pages. The scans originate from 212 pastoral districts (mainly parishes) with their own record keeping in the time between the uproar of 1848 and the beginning of the German Empire in 1871. This period is marked by massive social, economic and technical transformations.

According to the official order of the Catholic Church, the parish scribes had to record name, profession, religion, court, address, marital status, reason of death, dates of death and burial, age, names of doctor and priest as well as additional information in written form. The images display the records mainly in tabular format referring to one person per row. Stemming from more than 590 individual hands, the data set, recoding an estimated number of 295,000 dead persons, is also highly diverse with regards to writers.

A thorough analysis of the dataset shows that for 22,001 images 88 different table prints were used. These unique layouts were further categorized into eleven template categories. Most of these printed layout categories comply with the given normative for content imposed by the Church. The vast majority of scans (15,147 images) even fall into one single template category. On 4,578 pages, the requested information was recorded in manually drawn tables or manually extended table prints. The images are openly available through the *matricula* online platform[1], records can be queried using a search engine supplied by the Diocese of Passau[2]. The data are used by family historians as well as by historian scholars interested in age of those who died, the development or the spread of deadly diseases, etc.

The overall goal of this project is to perform Information Extraction on this collection, by extracting each death record and storing them in a database.

*B. Input Data and Workflow*

We consider the result of the following processing workflow as our input data for this experiments: starting from the image, graphical lines are first recognized using the algorithm described in [8]. Then the template matching tool is applied, which provides the column and header structure of the table. Text line detection and Hand-written Text Recognition (training included) tools are available through the Transkribus Platform [9]. Our input format is then similar to a traditional OCR output format: for a page, we have a set of regions defined by their position and size. We currently do no use textual information for the table understanding task.

*C. Template Processing*

The initial template-based table matching is done by analysing the separator lines, which define the table structure. Hand-drawn tables can show variations of the column width and row height (header). Additionally, variations can occur for the same table document, which has been used over years or decades in administration departments. Thus, a new table template matching that can deal with such variations has been developed. It matches the hierarchical structure of the table document and the defined template using an association graph (see Pelillo et al. [1] and Ishitano [2]) by finding a maximum clique. Thus, the columns and the defined header in the template are detected. Since in these registers, a double-page only contains one single table, this registration step also carries out the table localisation task (where is the table in the page).

*D. Our Main problem: Row Detection*

Once columns have been detected, the next task is to detect the Table Row, a challenging task for this collection for various reasons: First, separators (hand-drawn lines) are not used systematically for delimiting the rows (and are anyway recognized imperfectly). Then, the row layout depends on each writer and can vary inside one single record book. Some writers minimized the space between two rows, and, using a thin handwriting, also scribbled each record onto a single-line table row, while other writers made use of more space, and preferred centered lines, with some cells far longer than the others. Cells are also not always very well aligned horizontally, especially for these tables spreading over more than one page. Finally, the ditto sign (mostly denoted as '-') is often used for some columns (location), making the cell detection difficult. These issues are similar to those listed in [4].

Among other possibilities (see Section V), we choose to formulate the row detection problem as follows: Once the columns and the textlines have been identified, each textline will be tagged with one of the following categories: B, I, E, S, O, which correspond of the position of the textline in the cell (see Table 1):

TABLE 1: WE USE THE BIESO TAGSET FOR FORMULATING THE ROW DETECTION PROBLEM

| category | explanation |
|---|---|
| B(eginning) | First line of a cell |
| I(inside) | Line inside a cell (except first and last) |
| E(nd) | Last line of a cell |
| S(ingleton) | Single line of the cell |
| O(utside) | Outside a table |

This BIESO pattern is taken from the Natural Language Processing domain, and is used in order to recognize entities (sequence of words) in a sentence. Our assumption is that, once categorized, it will be easy to finally segment the table into rows. *Figure 1* shows an example of this annotation and Table 2 gives the label frequencies.

TABLE 2: FREQUENCIES OF THE DIFFERENT CLASSES IN OUR DATASET.

| **TextLine labels** | **Frequency** |
|---|---|
| B | 9947 |
| I | 9023 |
| E | 9941 |
| S | 8675 |
| O | 183 |
| **Total lines** | **37769** |

---

[1] http://data.matricula-online.eu/de/deutschland/passau/

[2] http://gendb.bistum-passau.de/

| Total cells | 18873 |

Once the BIESO categorization has been performed, the real final task is to build the rows (see Section III.B.3), and therefore obtain the cell structure, using the column information. Then Information Extraction can be applied on the table.

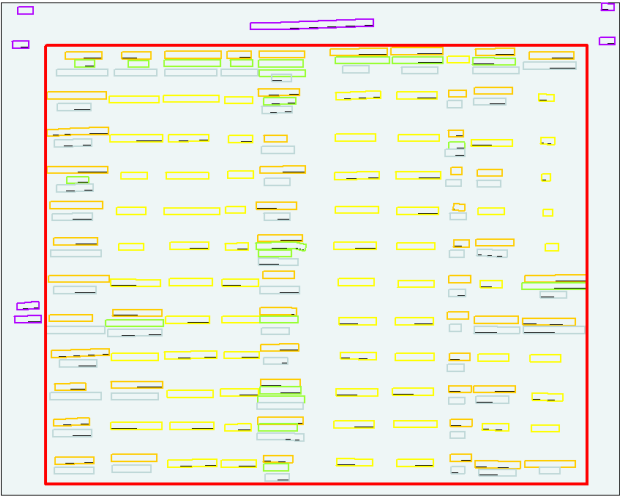

Figure 1. A table image. The BIESO labels on text lines are shown as orange, green, light blue, yellow and purple. The table zone is indicated in red.

## III. COMPARED APPROACHES

We describe here two approaches for solving our task: Graph-Conditional Random Fields and Graph and Edge Convolutional Networks. Both approaches rely on the same graph structure, which we explain first.

### A. Graph Structure

We model each page as a graph, where each node reflects one text line. An edge in the graph reflects a neighbouring relationship between two text lines, possibly long distance ones. More precisely, whenever there is horizontal, respectively vertical, significant and direct overlap between two bounding box of two text lines, we create a vertical, respectively horizontal, edge. 'Significant' means that the overlap must be higher than a certain threshold. 'Direct' means that the two bounding boxes must be in line of sight of each other, i.e. without any obstructing block in between.

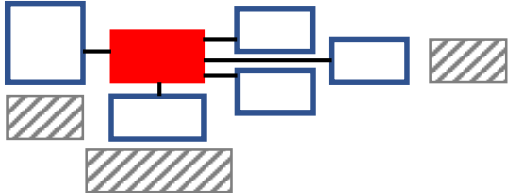

Figure 2: The neighbors of the red-filled block are in blue and are linked to it. Grey strikethrough blocks are not in its neighborhood.

This particular structure is arbitrary but based on the intuition that neighbouring relations are playing an important role. We have also experimented with acyclic structures based on minimum spanning tree but we observed a degradation of performance.

### B. Graph-CRF

The original CRF model [19] assumes that nodes are homogeneous, that is, they share the same label set. The prediction of a CRF model is multivariate, i.e. it is a vector of length $n$ of discrete labels in space $\{1, ..., l\}$.

$$Y = \{1, ..., l\}^n$$

Let $G = (V, E)$ where V denotes the set of vertices, and $E \subset V \times V$ the set of edges. The graph potential function $g(x, y)$ between an input $x$ and its structured label vector $y$ takes the form below.

$$g(x, y) = \sum_{v \in V} \psi_v(x, y_v) + \sum_{(v,w) \in E} \psi_{v,w}(x, y_v, y_w)$$

Here $\psi_v(x, y_v)$ is the unary potential function, while $\psi_{v,w}(x, y_v, y_w)$ is the pairwise potential functions.

Under the usual assumption that the potential functions are linear $g(x, y)$ is defined as:

$$g(x, y) = \sum_{v \in V} \theta_{y_v}^T \cdot \phi_V(v) + \sum_{(v,w) \in E} \vartheta_{y_v, y_w}^T \cdot \phi_E(v, w)$$

Here, $\theta_{y_v}$ is the model's unary weight vector given a node label $y_v$, $\phi_V(v)$ is a vector representation of the vertex $v$, $\vartheta_{y_v, y_w}$ is a pairwise weight vector given a pair of labels, and $\phi_E(v, w)$ is a vector representation of an edge $(v, w)$.

Training the CRF model involves choosing $\theta$ and $\phi$ that maximize $g(x, y)$ over the training set under some regularization.

Predicting, usually called inference, involves finding $y$ that maximizes $g(x, y)$ given some $x$.

### C. Edge Convolutional Networks

Graph Convolutional Networks (GCNs) [12] have been proposed recently for classifying nodes in a graph in a semi-supervised setting, i.e. when labels are only available for a subset of nodes. Although GCNs are standard feed-forward networks, there is one noticeable difference to standard networks which process elements independently from each other: GCNs operate on all the nodes of the graph at the same time and therefore introduce implicitly some dependencies among the predictions for unlabelled nodes. In other words, there is no notion of batch as the network takes the whole graph as input.

A GCN, in essence, is composed of two steps. First, compute some node representation by applying a transformation on the current feature representation; let us call this representation a potential in a loose sense. Then, this node potential is convolved: it simply means that for each node, one takes a weighted average of the neighbour nodes potential. Finally, this average is fed to the next layer of the neural network.

Let $A$ be the adjacency matrix, $X \in \mathbb{R}^{n,a}$ the features for the nodes, $H^i$ the layer in the GCN and $W_N^i$ nodes parameters at layer $i$. GCN models are then defined by

$$H^0 = X$$
$$H^i = f(D^{-0.5} A D^{0.5} H^{i-1} W_N^i)$$

where $f$ is any activation function (typically Relu), $A$ is the adjacency matrix and $D$ is a diagonal matrix of node degree. Looking closely at the above equation, the multiplication by $D^{-0.5}AD^{-0.5}$ amounts to taking a normalized average of the node representation.

In practice, Kipf and Welling [13] actually added a loop for each node, meaning that the intermediate representation (i.e. potential) for a node is actually a sum of its own potential and the potential of its neighbours. Finally, GCNs are trained with classic stochastic gradient descent such as Adam [15] and uses dropout to regularize the network [16].

One possible limitation of GCNs for our tasks could be the implicit assumption that nodes tend to have the same label as their neighbours, thus implicitly forming some *communities*. This hypothesis is recurrent in many works on social networks analysis and for some graph classification tasks. However, the graphs, reflecting the structure of documents, may contain many links between elements of different classes. Another significant limitation of GCNs, already mentioned in [13], is their inability to exploit edge features. Several related works on deep collective inference suffer also from the same limitations and employ Recurrent Neural Nets for node classification [10,11,12].

Therefore, we propose here a simple but new extension of GCNs to take into account edge features.

*1) Edges Convolutional Networks (ECNs)*

The main idea for ECNs is to learn graph convolutions which depends on edge features. If we can assign a score to each edge in the graph, we therefore have defined a parametrized adjacency matrix. In this way, the network can learn to filter out some edges or to find new ways to average neighbouring nodes. Let us consider the source edge matrix $S$ and target edge matrix $T$. $S_{ij}$ is 1 if edge $j$ has node $i$ as source (respectively destination for $T_{ij}$. Then,
$$A = S\,T^t$$
One way to define a parametrized adjacency matrix is:
$$g(w_E) = S\,diag(w_E\,F)T^t$$
where $F$ is the feature matrix for the edges in the graph and $w_E$ is a parameter vector for defining 1 convolution on edges.

Therefore, we could define Edge Graph Convolutional Networks by
$$H^i = f(\,g(w_E)\,H^{i-1}\,W_N^i)$$

The above model draws clearly inspiration from GCNs models, but instead of averaging the feature representation of neighbouring nodes, it uses the function $g$ to compute a weighted sum of neighbouring node depending on edge features. It means that thanks to the function $g$ and for instance Relu activation, the network is able to filter out some edges between nodes. Our proposal is seemingly related to the work of [14] where GCNs have been extended to model relational data.

**Variants:**

-*Stacking vs Adding*: Instead of adding the potential of neighbours to a node, one could simply stack it to the node. (this remove the self-loop). We could argue that it may not be a good idea to mix the representations of node and its neighbours, particularly if they have different labels. The network could therefore distinguish independently the labels from the neighbours from the node itself.

Note that when stacking, we implicitly change the size of the representation for the node. In our experiments, we have chosen the next layer parameter $W_N^i$ as a projection in a space with the same dimensionality of input nodes. (i.e. if a node has 10 features, stacking a single convolution gives another 10 features, so that the next layer $W_N^i$ reprojects in dimensionality 10).

-*Multiple Convolutions*. We could also consider multiple convolutions for each layer. In other words, we could consider different ways to average neighbouring nodes. In addition, theses convolutions would be learned by the system.

For instance, by stacking L different convolutions by layer, layer *I*, this Full Stacking variants is defined by

$$P^i = (\,H^{i-1}\,W_N^i)$$
$$H^i = f(P^i \oplus g(w_{E1})P^i \oplus \ldots \oplus g(w_{EL})P^i\ ),$$

where $\oplus$ denotes the concatenation operator.

Another variant, that does not mix the node representation with the one from its neighbours, is to simply average the results of the convolutions. The Sum Stacking is therefore defined by

$$H^i = f(P^i \oplus \frac{1}{L}\sum_{c=1}^{L} g(w_{Ec})P^i)$$

IV. EXPERIMENTS

Finally, we evaluated the CRF and GCN models both on the BIESO task, but also on the Table Row detection task using the workflow designed *per se.* The code and the datasets are available from [25] .

*A. Datasets*

*1) Dataset1*

For training and evaluating the algorithms, our dataset (hereafter named as dataset1) is composed of 144 manually annotated pages (which amounts for 37,769 textlines or 18,873 cells). Training and testing have been performed using a 4-fold cross-validation. The full dataset (image and pickled extracted features), as well as the 4-fold repartition are available in the usecases/ABP folder of [25].

*2) Dataset2*

A second 150-page dataset (dataset2) from 15 different books, representative of the ABP_S_1847-1878 dataset (cf. Section II.A) is also used in order to access the overall Table Understanding. For this the IE workflow is used to exploit textlines detection, template registration and row detection.

*B. Experiments and Evaluations*

This work is open source and available from [25] (See the file tasks.DU_ABPTable_Quantile.py).

*1) Features*

For all models, the node features consist of geometrical features, e.g. width, height of bounding boxes, etc. For models able to exploit edge features, the latter characterize the geometric neighbouring relation, e.g. distance or justification, etc. There 29 node features and 140 edge features.

*2) Logistic regression*

In order to show the importance of using structured machine learning, we also experimented with classifying each node independently from the others. We therefore trained 2 node classifiers using a logit model.

The first model, called *Logit-Standard*, exploits the feature of the node.

For the second model, called *Logit-1conv*, each node representation is built by concatenating the node's features with a single convolution (weighted sum of the neighbors' node features). This is a simple trick to provide some context information to the "flat" classifier.

*3) CRF setup*

We implemented our CRF model using the Open Source Python library called PyStruct [20], more precisely an open source extension of it [21][23]. We trained using the one-slack structured SVM method and run inferences using AD3. We trained with 1500 iterations using the hyper-parameter default values.

*4) GCN and ECN setup*

First, we implemented some standard GCN models, i.e. without edge features. We experimented with several architectures but GCNs were not able to fit the training set, with a test performance around 70%. In the next table, we report the performance of GCN with 5 layers, using a stacking approach for the convolution result. As previously mentioned, it could be explained by using a single convolution, their inability to take into account edge features but also by the community hypothesis.

We then experimented with different versions of ECNs, by varying the number of layers and the number of convolutions. For brevity's sake, we do not report all the architectures we have tested. However, our observations showed us that that the number of parameters need to be high enough to be able to fit the training set.

In practice, we used 10% of the training set as a validation set for model selection. In practice, we found that a model with 3 layers and 10 convolutions by layers performed reasonably well. We trained the model for at most 2,000 epochs and with a learning rate of 0.001 with Adam and use the validation set for early stopping.

*5) BIESO evaluation*

Table 3 compares the classification accuracy of Logit, GCN, CRF and ECN models on a 4-fold experiment. In short, these results indicate that the new models, ECNs, are effective and outperform standard GCNs and Logit baselines. In addition, ECNs obtain similar performances as the CRF ones, even if they rely on very different inference algorithms. ECNs only employ matrix multiplications and non-linear activations, while CRFs rely on pairwise potentials and a structured SVM method.

*6) Table Row Evaluation*

In this setting, we evaluate the real workflow: all the steps are automatically performed: template registration for column segmentation and textline detection. Then, this input is sent to our Row detector (first BIESO categorization, then row construction), and a final evaluation at row level is computed.

Once textlines are categorised with the BIESO, we segment each column independently into cells using the BIESO tags. At this stage, some inconsistencies across columns may occur due to BIESO tagging errors. Then the next step is to globally select the horizontal cuts (Y-cuts) which will define the table rows. For this, we select as Y-cuts candidates the top Y of each cell, and cluster them (agglomerative cluster with textline height as stop criterion). Afterwards, a cluster is considered as Y-cut if its number of elements (one per column) is greater than a proportion of the total number of columns (in practice 0.33). This is far from been optimal (cf. Section V.), but works well enough for the Information Extraction tool (assuming a perfect BIESO tagging, this method reaches a precision of 0.916 and a recall of 0.967). Especially skewing is not supported. Evaluation is performed using the precision/recall model as defined by [24] (as correct and missed detection) with an overlap threshold of TH=0.50 (we tried various values, without any impact on the comparison). For this evaluation, both CRF and ECN were trained with the full dataset1.

TABLE 3: ACCURACY FOR BIESO TAGGING (DATASET1)

| Method | #params | Fold 1 | Fold 2 | Fold 3 | Fold 4 | Avg |
|---|---|---|---|---|---|---|
| Logit-standard | 150 | 0.37 | 0.37 | 0.36 | 0.36 | 0.37 |
| Logit-1conv | 430 | 0.39 | 0.39 | 0.65 | 0.61 | 0.51 |
| GCN | 7893 | 0.76 | 0.75 | 0.76 | 0.70 | 0.74 |
| CRF (1500 iter.) | 3645 | 0.95 | 0.92 | 0.92 | 0.89 | 0.92 |
| 3Layer-10conv-FullStack | 25172 | **0.96** | **0.94** | **0.93** | 0.92 | **0.94** |
| 8Layer-1Conv | 14059 | 0.95 | **0.94** | 0.92 | 0.91 | 0.93 |

*C. Discussion*

Table 4 shows that finally the CRF approach is slightly better than ECN. For this second dataset, which we considered as easier, ECN losses 5 points in accuracy, while CRF losses only 1 point. Overfitting could be an explanation. Some models with early stopping or drop-out do not show any improvements. Note that in Table 4, the BIEOS accuracy is computed using the manually annotated dataset (textlines and cells), while Precision Recall, and F-1 score are computed using the automatic workflow where textlines were detected. We also tested a degraded version of CRF with only 100 iterations. As expected the BIESO accuracy decreases significantly (-15.7), while at the row detection evaluation, F-1 only losses 5.5 points. Its seems that the way final rows are built has a strong impact on the final quality.

TABLE 4: EVALUATION (PRECISION, RECALL, F-1 AND BIESO ACCURACY) FOR TABLE ROW DETECTION (DATASET2)

| Method | P | R | F-1 | BIESO ACC |
|---|---|---|---|---|
| CRF (1500 iterations) | 0.864 | 0.933 | 0.897 | 0.911 |
| CRF (100 iterations) | 0.785 | 0.907 | 0.842 | 0.754 |
| 3Layer-10conv-FullStack | 0.842 | 0.930 | 0.884 | 0.881 |
| 8Layer-1Conv | 0.856 | 0.932 | 0.892 | 0.890 |

We also tested 'non-structural' approaches such as logistic regression, which totally failed (first two rows in Table 4) .

In order to assess the importance of edge features compared to node features, we also experimented with discarding node or edges features. This can be simply done by representing each node (resp. edge) by a 1-dimension vector containing a 1. Discarding edge features leads to a low accuracy of 0.33. Differentiating vertical edges from horizontal ones without other edge features increases this accuracy up to 0.50. Finally, when discarding node features, accuracy drops by less than 1%, which shows the importance of the edge features, the node features being of no real use to CRF. Of course, if we had some textual features for nodes, then we would expect a larger difference.

In term of computation time, ECN outperforms CRF even if ECN have more parameters. Using similar CPUs and relatively equivalent RAM, ECN is training about 4 time faster, and predicting about 40 time faster than CRF. ECN also scales up better than CRF since CRF must load in memory the whole dataset for performing the structured SVM algorithm while the neural network is trained by batch of data. Finally, ECN can easily be ran on GPUs.

Noticeably, it is worth mentioning that a key component of the IE workflow, the textline detector, shows a great robustness, since the full table structuring task relies on textlines. Also some evaluation has been conducted for the IE workflow with 1,000 pages, and shows that the row detection performs similarly to the evaluation datasets presented here.

## V. Conclusion and Future Work

While the current workflow is efficient enough for performing the Information Extraction task, whatever the method used, we still aim at improving the Table Understanding task. The next short-term step is to better perform the Table Row detection, which is crucial as records are organised per row. As said, the current implementation was quickly designed so that this functionality was available for the full IE workflow.

Another direction is to create a more diverse dataset, from various providers in order to access the generalisation power of both models. As we have shown that ECNs are competitive to CRF-based model, we would like to further extend these models to explore other Document Understanding tasks. A longer-term step is to generalise the method to encompass the full Table Understanding task for handwritten, printed and digital-born documents: These experiments, among others, show that a key component, the textline detector, works very well for the handwritten document. Assuming a common input for all types of documents may be realistic, so that a common solution can be designed for the three types of documents. While an appropriate dataset for handwritten documents has still to be annotated (by enriching the current one and by adding documents from various sources), we first foresee to use the ICDAR 2013 dataset (printed/digital-born documents) to asses this hypothesis.

### Acknowledgment

This project has received funding from the European Union's Horizon 2020 research and innovation programme under grant agreement No 674943 (READ project).